\documentclass{article}
\usepackage[preprint]{colm2025_conference}

\usepackage{microtype}
\usepackage[utf8]{inputenc} 
\usepackage[T1]{fontenc}    
\definecolor{green}{RGB}{0,150,10}
\definecolor{blue}{RGB}{0,148,181}
\definecolor{orange}{RGB}{194,153,107}
\usepackage[colorlinks,linkcolor=red,anchorcolor=green,citecolor=blue]{hyperref}
\usepackage{lineno}
\usepackage{booktabs}       
\usepackage{tabularx}
\usepackage{amsfonts}       
\usepackage{nicefrac}       
\usepackage{xcolor}         
\usepackage{amsmath}
\usepackage{marvosym}
\usepackage{graphicx}
\usepackage{CJKutf8}
\usepackage{colortbl}
\usepackage{multirow}
\usepackage{subcaption}
\usepackage{changes}
\usepackage{listings}
\usepackage{mdframed}
\usepackage{caption}
\usepackage{colortbl}
\usepackage{multicol}
\usepackage{arydshln}

\usepackage[capitalize,noabbrev]{cleveref}

\title{MolAct: An Agentic RL Framework for Molecular Editing and Property Optimization}

\author{
Zhuo Yang$^{1,2}$\footnotemark[1] \And Yeyun Chen$^{1,3}$\footnotemark[1] \And Jiaqing Xie$^{1}$ \And Ben Gao$^{1,4}$ \And Shuaike Shen$^{8}$ \AND Wanhao Liu$^{1}$ \And Liujia Yang$^{1,5}$ \And Beilun Wang$^{6}$ \And Tianfan Fu$^{7,1}$ \And Yuqiang Li$^{1}$\footnotemark[2] \\[3mm]
$^{1}$Shanghai Artificial Intelligence Laboratory \\
$^{2}$Xidian University \quad
$^{3}$Shanghai Innovation Institute \quad
$^{4}$Wuhan University \\
$^{5}$Shanghai Jiao Tong University \quad
$^{6}$Southeast University \quad
$^{7}$Nanjing University \\
$^{8}$Carnegie Mellon University \\[1.5mm]
{\small Github: \url{https://github.com/little1d/MolAct}} \\
{\small Huggingface: \url{https://huggingface.co/collections/little1d/molact}}
}

\begin{document}
\begin{CJK*}{UTF8}{gbsn}

\maketitle
\footnotetext[1]{Equal contribution.}
\footnotetext[2]{Corresponding authors.}

\begin{figure*}[h]
    \centering
    \begin{subfigure}{0.48\textwidth}
        \centering
        \includegraphics[width=\linewidth]{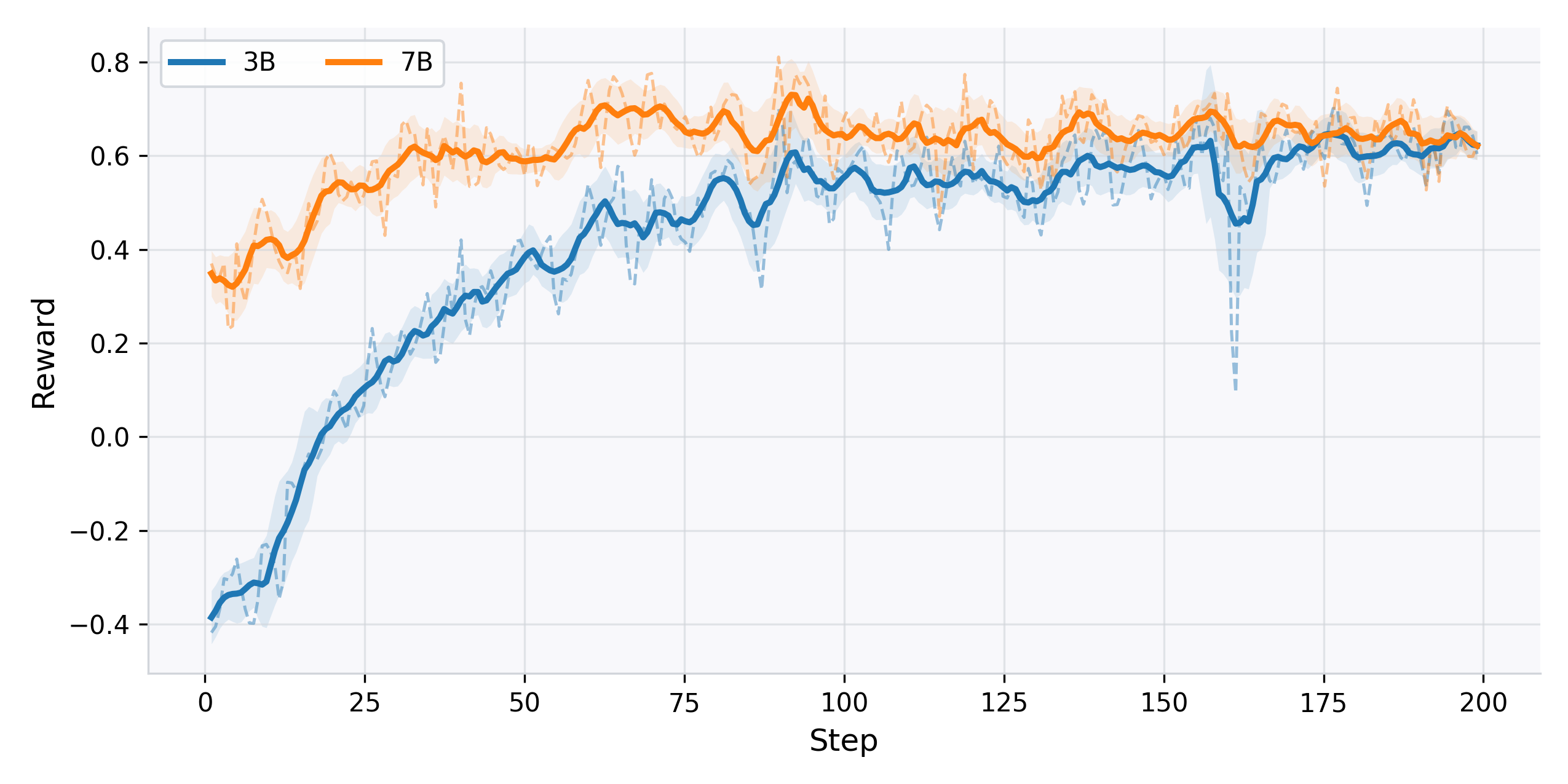}
        \caption{\textbf{Stage 1: Molecular Edit Tasks}}
        \label{fig:front_edit_curve}
    \end{subfigure}
    \hfill
    \begin{subfigure}{0.48\textwidth}
        \centering
        \includegraphics[width=\linewidth]{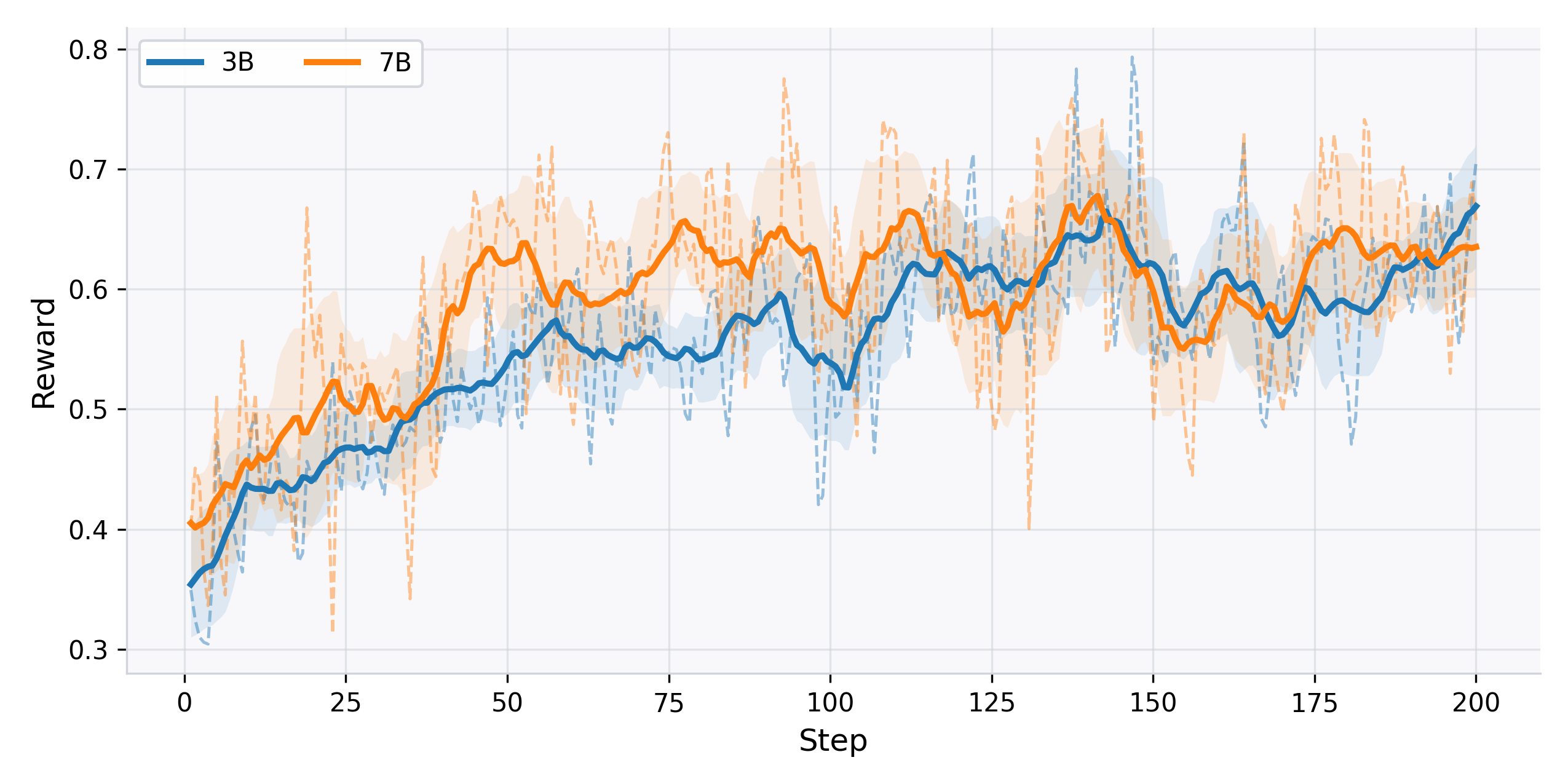}
        \caption{\textbf{Stage 2: Molecular Optimization Tasks}}
        \label{fig:front_opt_curve}
    \end{subfigure}
    \caption{Training reward curves comparing Qwen-2.5-3B and Qwen-2.5-7B backbones for molecular editing (Stage 1) and molecular optimization (Stage 2, continued training from Stage 1). On simpler molecular editing tasks, the 7B model converges faster and achieves higher performance compared to the 3B model. Additionally, in the more challenging molecular optimization tasks, the 7B model continues to demonstrate superior performance.}
    \label{fig:front_reward_combo}
\end{figure*}

\begin{abstract}
    Molecular editing and optimization are multi-step problems that require iteratively improving properties while keeping molecules chemically valid and structurally similar. We frame both tasks as sequential, tool-guided decisions and introduce \textbf{MolAct}, an agentic reinforcement learning framework that employs a two-stage training paradigm: first building editing capability, then optimizing properties while reusing the learned editing behaviors. To the best of our knowledge, this is the first work to formalize molecular design as an Agentic Reinforcement Learning problem, where an LLM agent learns to interleave reasoning, tool-use, and molecular optimization. The framework enables agents to interact in multiple turns, invoking chemical tools for validity checking, property assessment, and similarity control, and leverages their feedback to refine subsequent edits. We instantiate the MolAct framework to train two model families: \textbf{MolEditAgent} for molecular editing tasks and \textbf{MolOptAgent} for molecular optimization tasks. In molecular editing, \textbf{MolEditAgent-7B} delivers 100, 95, and 98 valid add, delete, and substitute edits, outperforming strong closed "thinking" baselines such as DeepSeek-R1; \textbf{MolEditAgent-3B} approaches the performance of much larger open "thinking" models like Qwen3-32B-think. In molecular optimization, \textbf{MolOptAgent-7B} (trained on MolEditAgent-7B) surpasses the best closed "thinking" baseline (e.g., Claude 3.7) on LogP and remains competitive on solubility, while maintaining balanced performance across other objectives. These results highlight that treating molecular design as a multi-step, tool-augmented process is key to reliable and interpretable improvements.
\end{abstract}

\footnotetext[3]{A famous Chinese proverb. English translation: ``To do a good job, one must first sharpen one's tools.'' Confucius (551--479 BCE) was a Chinese philosopher and teacher, also known as Kongzi (孔子).}

\begin{quote}
\textcolor{gray}{``工欲善其事，必先利其器.''}\footnotemark[3] --- Confucius
\end{quote}

\section{Introduction}



Molecular editing and molecular optimization are fundamental problems in computer-aided drug discovery and molecular design \citep{Gomez-Bombarelli2018-ly, Ma2024-ys, Landrum2025-vh}. In practice, these tasks involve modifying a given molecule through a series of structural operations—such as adding, deleting, or substituting functional groups—in order to improve physicochemical properties or biological activities while preserving chemical validity and structural similarity to the original compound \citep{DBLP:journals/corr/abs-1802-04364, Brown_2019}. Effective molecular editing and optimization inherently require careful coordination between local structural changes and global property objectives, a characteristic long recognized in medicinal chemistry workflows \citep{yang2021improvingmoleculardesignstochastic, Han2024-gk}.


A wide range of computational approaches have been proposed to address molecular editing and optimization, including graph-based generative models \citep{Jensen2019-vj, Erikawa2023-cl}, variational autoencoders \citep{DBLP:journals/corr/abs-1802-04364,DBLP:journals/corr/abs-1802-03480, Nguyen2025-dg}, and reinforcement learning methods \citep{zhuang2025moleditrlstructurepreservingmolecularediting, Haddad2025-rg, Lin2025-ju}. These methods have achieved notable progress in optimizing specific molecular properties or generating chemically valid structures under predefined constraints. However, many of them rely on rigid model assumptions, domain-specific heuristics, or fixed optimization pipelines, which limit their flexibility when adapting to diverse objectives or complex structural constraints \citep{yong2025bayesianoptimizationmoleculesparetoaware}. More recently, large language models (LLMs) have been introduced into molecular modeling and design tasks, leveraging their strong representation learning and instruction-following capabilities \citep{ye2023drugassistlargelanguagemodel, dey2025gellmogeneralizinglargelanguage}. Despite promising results, most LLM-based approaches still operate in a static generation or instruction-tuning paradigm, producing modified molecules in a single pass without explicitly modeling the sequential decision process underlying molecular editing and optimization. As a consequence, these methods struggle to capture the step-by-step dependency between intermediate molecular modifications and final property outcomes, and they lack an explicit mechanism to incorporate structured feedback from molecular validity checks or property evaluations during the optimization process. To the best of our knowledge, this is the first work to formalize molecular design as an Agentic Reinforcement Learning problem, where an LLM agent learns to interleave reasoning, tool-use, and molecular optimization.

From a decision-making perspective, molecular editing and optimization involve sequences of interdependent choices, where each modification influences downstream outcomes. Ensuring chemical validity, maintaining scaffold similarity, and accurately assessing property changes depend on external chemical tools \citep{Landrum2025-vh, huang2021therapeuticsdatacommonsmachine}. Without explicit tool-based feedback, models may hallucinate chemically invalid structures that violate valence rules or invent impossible functional groups \citep{Gao2020-ak, Polykovskiy2020-ox, BENDER2021511}. In practice, chemists rely on immediate checks—validity, property estimators, scaffold similarity—to prune failures early, making tool-based feedback indispensable for reliable molecular design.

To facilitate systematic evaluation of molecular editing and optimization methods, ChemCoTBench \citep{li2025chemicalqaevaluatingllms} provides a collection of well-defined tasks that cover both elementary molecular editing operations and property-guided molecular optimization objectives. The benchmark includes functional group addition, deletion, and substitution tasks, as well as optimization targets spanning physicochemical properties and protein-related bioactivities. By offering standardized task definitions, curated datasets, and quantitative evaluation metrics, ChemCoTBench enables consistent comparison across different modeling approaches and optimization strategies. In this work, we adopt the molecular editing and molecular optimization tasks from ChemCoTBench as our primary experimental benchmark, using them to assess the effectiveness of our approach across diverse molecule modification scenarios and property objectives.

In this paper, we propose \textbf{MolAct}, an agentic reinforcement learning framework that models molecule modification as a sequential, tool-augmented decision process. We instantiate MolAct to train \textbf{MolEditAgent} and \textbf{MolOptAgent} for editing and optimization tasks, respectively. Our experiments on ChemCoTBench demonstrate the effectiveness of a two-stage training paradigm that first builds editing capabilities before tackling property optimization.

Our main contributions can be summarized as follows:
\begin{itemize}
    \item We formulate molecular editing and molecular optimization as multi-step decision-making problems with explicit verification and feedback, highlighting the importance of sequential structure modification in molecular design.
    \item We present MolAct, a tool-augmented agent framework that integrates external chemical evaluations to guide molecule modification.
    \item We systematically evaluate MolAct on ChemCoTBench\citep{li2025chemicalqaevaluatingllms} across diverse editing and optimization tasks, achieving strong validity and property outcomes with a compact two-stage, multi-turn RL approach.
\end{itemize}

\section{Problem Formulation}
We cast molecular editing and molecular optimization as sequential, tool-mediated decisions over a vast, discrete chemical space. Practical drug design starts from a lead scaffold and relies on iterative edits with rapid feedback; a one-shot formulation cannot capture this workflow.

\subsection{Tasks and State Space}
\textbf{State.} A state $s_t$ is the current molecule encoded as a SMILES string. Edits are applied to SMILES with pattern-based operators, and chemical validity is enforced by external checks.

\textbf{Molecular editing.} Given $s_{\text{src}}$ and an edit instruction, return $s_{\text{edit}}$ that (i) applies the instructed change, (ii) is chemically valid, and (iii) remains similar to $s_{\text{src}}$. The actionable edit operators are \textit{add}, \textit{delete}, and \textit{substitute} functional groups.

\textbf{Molecular optimization.} Given $s_{\text{src}}$ and a target objective, the goal is to generate $s_{\text{opt}}$ that achieves improved performance on the specified objective while maintaining both chemical validity and structural consistency with the source molecule. In our formulation, structural consistency is specifically enforced at the level of the Murcko scaffold \citep{Bemis1996-hu}. The considered objectives encompass physicochemical properties such as LogP, solubility, and QED, as well as biologically relevant activities including DRD2, JNK3, and GSK3$\beta$.

\subsection{Actions and Transitions}
At step $t$, the policy $\pi(a_t \mid s_t)$ chooses: (i) an edit operator with a tool-validated attachment/removal site, yielding a new molecule $g(s_t,a_t)$; (ii) a tool/evaluator call (validity, similarity, property oracle) that returns feedback without changing the molecule; or (iii) a terminate action. We define the transition function
\[
s_{t+1} = f(s_t, a_t) =
\begin{cases}
g(s_t,a_t), & \text{if } a_t \text{ is an edit},\\
s_t, & \text{if } a_t \text{ is an evaluation call},\\
s_t, & \text{if } a_t \text{ is terminate},
\end{cases}
\]
and the trajectory $\tau=(s_0,a_0,o_0,s_1,a_1,o_1,\dots,s_T)$ starts at $s_0=s_{\text{src}}$ and ends on terminate or a turn budget, where each $o_t$ is the tool observation returned by the chosen action (edit or evaluator).

\subsection{Rewards and Objectives}
For a trajectory $\tau$, we maximize the expected return $\mathbb{E}_{\pi}[\sum_{t=0}^T r_t]$. Final rewards (detailed in Method) combine: a validity gate (invalid $\to$ fixed penalty), task success (edit correctness or property improvement), structural similarity or scaffold preservation, and a small bonus for grounded tool use. Rewards are bounded for stability and applied only to agent tokens; tool outputs are context only. Editing rewards prioritize correct application of the operator and similarity to $s_{\text{src}}$; optimization rewards prioritize property gain and scaffold retention.

\subsection{Why Multi-Step Control}
The molecular space is astronomically large and ill-behaved for direct search. A multi-step, tool-guided process lets the policy decompose large moves into smaller, verifiable edits, leverage intermediate feedback to avoid invalid regions, and operate under a tight interaction budget—mirroring real medicinal chemistry practice and motivating the two-stage training that first stabilizes editing before property-driven optimization.

\section{Results}
\subsection{Overview of the MolAct Framework}

\begin{figure}[!h]
    \centering
    \includegraphics[width=\textwidth]{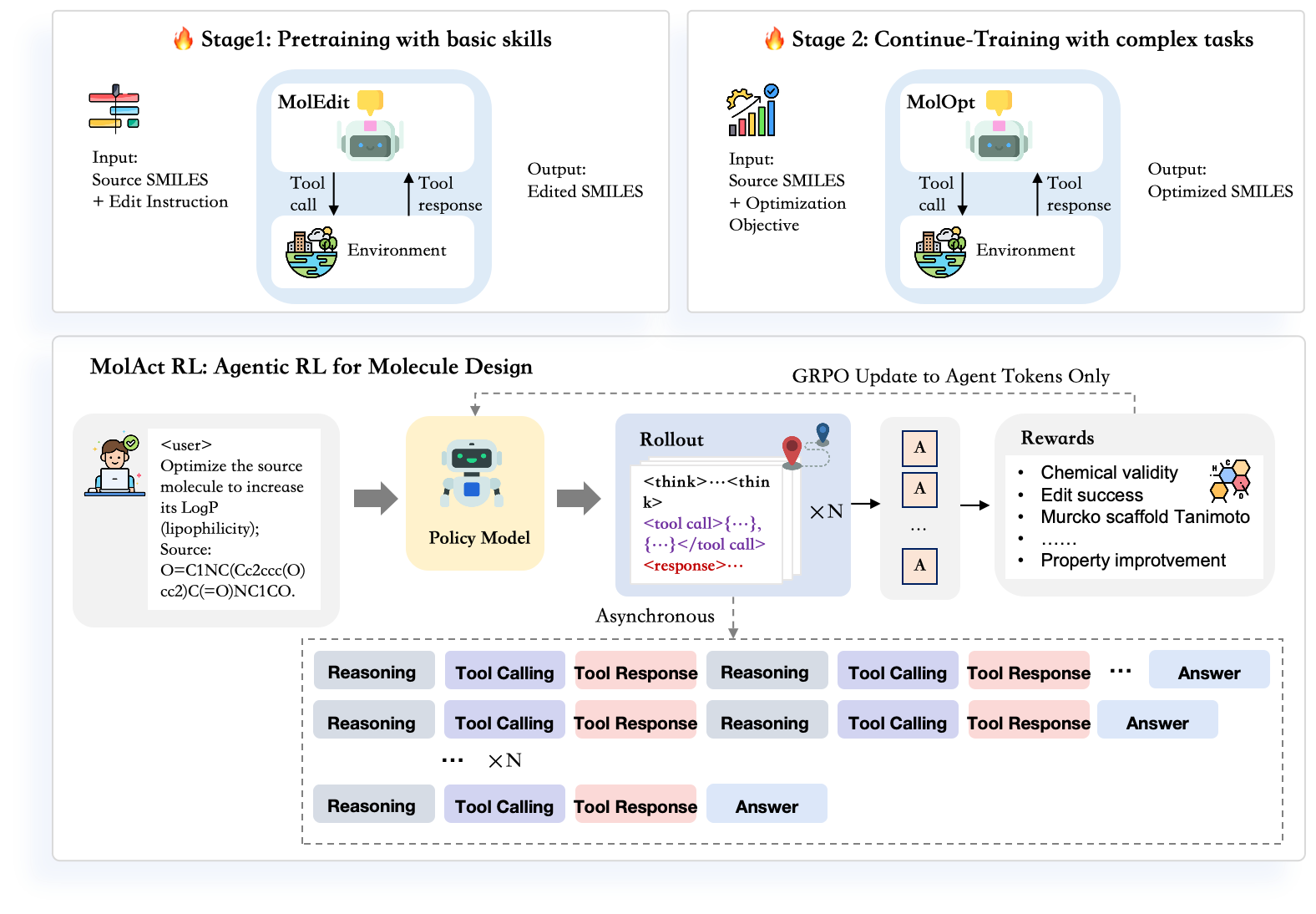}
    \caption{MolAct overview. Two-stage training: Stage 1 pretraining with basic edit skills, Stage 2 continue-training on complex optimization tasks, using multi-turn tool calls, group-relative rollouts, and masked updates on agent tokens.}
    \label{fig:molact_framework}
\end{figure}

\textbf{Terminology and Framework Structure.} \textbf{MolAct} is the agentic reinforcement learning framework that provides the training methodology and tool-augmented decision process. We instantiate MolAct to train two model families: \textbf{MolEditAgent} for molecular editing tasks and \textbf{MolOptAgent} for molecular optimization tasks. Each model family contains multiple model variants based on different backbone sizes: for example, \textbf{MolEditAgent-7B} and \textbf{MolEditAgent-3B} refer to MolEditAgent models trained on Qwen-2.5-7B and Qwen-2.5-3B backbones, respectively; similarly, \textbf{MolOptAgent-7B} and \textbf{MolOptAgent-3B} are MolOptAgent variants based on 7B and 3B backbones. Throughout this paper, we use the family names (MolEditAgent, MolOptAgent) when discussing general capabilities or training procedures, and the specific variant names (e.g., MolEditAgent-7B) when reporting experimental results for particular model instances.

Figure~\ref{fig:molact_framework} summarizes MolAct: a two-stage agentic RL framework. Stage 1 pretraining (\textbf{MolEditAgent}) learns basic edit primitives (add, delete, substitute) with validity and similarity feedback. Stage 2 continue-training (\textbf{MolOptAgent}) initializes from Stage 1 and learns complex property optimization under validity and scaffold constraints. For each prompt, we copy it into $K$ parallel rollout chains (group-relative), run multi-turn "think $\rightarrow$ tool call $\rightarrow$ observation" cycles until terminate or a turn budget, and apply masked updates only on agent tokens. Rewards combine validity, task success (edit correctness or property gain), similarity/scaffold, and a tool-use bonus. At inference, the same loop selects a valid SMILES that meets the task objective within the budget.

\subsection{Experimental Setup}

\paragraph{Dataset.}
We construct our training dataset from ChemCoTDatasets \citep{li2025chemicalqaevaluatingllms}, which contains annotated chain-of-thought data for multi-step molecule modification.
We remove the chain-of-thought reasoning process and extract the questions (source molecules with task specifications) and reference answers (target molecules) to form our reinforcement learning dataset.
Each training instance consists of a source molecule (SMILES string) paired with either an editing instruction (for Stage 1) or a target optimization objective (for Stage 2).
All molecules are canonicalized and validated to ensure consistent input representations.

\paragraph{Task Settings.}
For molecular editing, we evaluate functional group addition, deletion, and substitution tasks.
For molecular optimization, we consider six optimization objectives, including LogP, solubility, QED,
and predicted bioactivities against DRD2, JNK3, and GSK3$\beta$.

\paragraph{Evaluation Metrics.}
For molecular editing tasks, we use Pass@1 to assess whether the edited molecule meets the editing instructions, and also report molecule validity to ensure chemical correctness.
For molecular optimization tasks, we report two metrics following the benchmark protocol: $\Delta$ (mean property improvement, where a negative $\Delta$ indicates that most optimizations result in property degradations) and SR\% (success rate, the percentage of optimizations that achieve property increase).

\subsection{Main Results}

Figure~\ref{fig:front_reward_combo} shows the training reward curves for both stages.

\paragraph{Molecular Editing.}
Table~\ref{tab:acc_performance} presents the accuracy performance of various large language models on molecular editing tasks within ChemCotBench. \textbf{MolEditAgent-7B} achieves 90.0\% for Add, 80.0\% for Delete, and 78.3\% for Sub operations. Notably, while Gemini-2.5-pro-think leads with 100.0\% on Add, MolEditAgent-7B outperforms all W/o Thinking models (best: GPT-4o at 80.0\%) and ranks second overall. For Delete and Sub tasks, MolEditAgent-7B matches or exceeds most baselines, including several W/ Thinking models.

\begin{table}[!h]
    \centering
    \caption{Accuracy (Acc.) Performance Comparison on Molecular Editing Tasks (\%). Benchmarked results are quoted from \citep{li2025chemicalqaevaluatingllms}.}
    \label{tab:acc_performance}
    \begin{tabular}{lccc}
        \toprule
        \textbf{Models} & \textbf{Add (Acc.)} & \textbf{Delete (Acc.)} & \textbf{Sub (Acc.)} \\
        \midrule
        \multicolumn{4}{l}{\textit{W/ Thinking}} \\
        \hdashline
        Gemini-2.5-pro-think & 100.0 & 85.0 & 81.7 \\
        Claude3.7-sonnet-think & 85.0 & 80.0 & 83.4 \\
        DeepSeek-R1 & 70.0 & 70.0 & 68.3 \\
        o3-mini@20250103 & 65.0 & 55.0 & 80.0 \\
        o1-mini@20240912 & 55.0 & 80.0 & 58.3 \\
        Qwen3-235B-A22B-think & 40.0 & 75.0 & 71.7 \\
        Qwen3-32B-think & 20.0 & 55.0 & 20.0 \\
        Llama-Nemo-49B-think & 0.0 & 80.0 & 8.0 \\
        \midrule
        \multicolumn{4}{l}{\textit{W/o Thinking}} \\
        \hdashline
        GPT-4o@20241120 & 80.0 & 80.0 & 65.0 \\
        Deepseek-V3 & 70.0 & 75.0 & 76.7 \\
        Gemini-2.0-flash & 65.0 & 75.0 & 66.7 \\
        Qwen2.5-235B-A22B & 40.0 & 75.0 & 66.7 \\
        Qwen3-32B & 30.0 & 55.0 & 25.0 \\
        Qwen2.5-72B-Instruct & 70.0 & 80.0 & 56.7 \\
        Qwen2.5-32B-Instruct & 50.0 & 50.0 & 48.3 \\
        Llama-3.1-70B-Instruct & 60.0 & 80.0 & 50.0 \\
        Llama-Nemo-49B & 30.0 & 55.0 & 30.5 \\
        Gemma-2-27b-it & 75.0 & 70.0 & 35.0 \\
        Phi-4-14B & 60.0 & 80.0 & 38.3 \\
        OLMo2-32B-Instruct & 15.0 & 30.0 & 11.7 \\
        BioMedGPT-7B & 10.0 & 12.0 & 10.0 \\
        BioMistral-7B & 0.0 & 10.0 & 0.0 \\
        \midrule
        \textbf{MolEditAgent-7B} & \textbf{90.0} & \textbf{80.0} & \textbf{78.3} \\
        \textbf{MolEditAgent-3B} & \textbf{80.0} & \textbf{70.0} & \textbf{16.7} \\
        \bottomrule
    \end{tabular}
\end{table}

\noindent
\textbf{MolEditAgent-3B} achieves 80.0\% for Add and 70.0\% for Delete, demonstrating scalability with model capacity. However, its Sub accuracy (16.7\%) is notably lower, likely due to substitution requiring more complex multi-step reasoning that benefits from larger models. Table~\ref{tab:valid_improvement} further shows that MolAct models maintain high molecular validity (95.0--100.0\% for 7B, 71.7--95.0\% for 3B), substantially outperforming base instruction-tuned models (Qwen2.5-7B: 65.0--75.0\%, Qwen-2.5-3B: 55.0--65.0\%), confirming that agentic RL training improves both accuracy and chemical validity.

\begin{table}[htbp]
    \centering
    \caption{Validity Improvement from MolAct Framework (\%)}
    \label{tab:valid_improvement}
    \begin{tabular}{l ccc}
        \toprule
        \textbf{Models} & \textbf{Add (Valid.)} & \textbf{Delete (Valid.)} & \textbf{Sub (Valid.)} \\
        \midrule
        \textbf{MolEditAgent-7B} & \textbf{100.0} & \textbf{95.0} & \textbf{98.0} \\
        Qwen2.5-7B-Instruct & 75.0 & 70.0 & 65.0 \\
        \midrule
        \textbf{MolEditAgent-3B} & \textbf{95.0} & \textbf{80.0} & \textbf{71.7} \\
        Qwen-2.5-3B-Instruct & 60.0 & 55.0 & 65.0 \\
        \bottomrule
    \end{tabular}
\end{table}

\paragraph{Molecular Optimization.}
Table~\ref{tab:molopt_performance} presents results on six molecular optimization tasks. \textbf{MolOptAgent-7B} (pre-trained on MolEditAgent-7B) achieves the highest LogP $\Delta$ of \textbf{0.89} (SR\% \textbf{92}), outperforming all baselines including Claude3.7-sonnet-think (0.41) and Gemini-2.0-flash (0.35). For Solubility, it achieves $\Delta$ \textbf{1.42} (SR\% \textbf{84}), second only to Gemini-2.5-pro-think (1.91) and DeepSeek-R1 (1.48) in $\Delta$, while maintaining competitive SR\%. On bioactivity targets, MolOptAgent-7B shows positive improvements: DRD2 $\Delta$ \textbf{0.02} (SR\% \textbf{38}) and GSK-3$\beta$ $\Delta$ \textbf{0.04} (SR\% \textbf{36}), though lower than Gemini-2.5-pro-think's DRD2 $\Delta$ 0.35. QED ($\Delta$ \textbf{0.04}, SR\% \textbf{35}) and JNK3 ($\Delta$ \textbf{-0.04}, SR\% \textbf{14}) show modest results, indicating room for improvement on these objectives.

\begin{table}[!h]
    \centering
    \caption{Performance Comparison on Molecular Optimization Tasks. Higher $\Delta$ and SR\% are better. Benchmarked results are quoted from \citep{li2025chemicalqaevaluatingllms}.}
    \label{tab:molopt_performance}
    \resizebox{\textwidth}{!}{
    \begin{tabular}{l|rr|rr|rr|rr|rr|rr}
        \toprule
        \textbf{Models} & \multicolumn{2}{c|}{\textbf{LogP}} & \multicolumn{2}{c|}{\textbf{Solubility}} & \multicolumn{2}{c|}{\textbf{QED}} & \multicolumn{2}{c|}{\textbf{DRD2}} & \multicolumn{2}{c|}{\textbf{JNK3}} & \multicolumn{2}{c}{\textbf{GSK-3$\beta$}} \\
        \cmidrule(lr){2-3} \cmidrule(lr){4-5} \cmidrule(lr){6-7} \cmidrule(lr){8-9} \cmidrule(lr){10-11} \cmidrule(lr){12-13}
        & \textbf{$\Delta$} & \textbf{SR\%} & \textbf{$\Delta$} & \textbf{SR\%} & \textbf{$\Delta$} & \textbf{SR\%} & \textbf{$\Delta$} & \textbf{SR\%} & \textbf{$\Delta$} & \textbf{SR\%} & \textbf{$\Delta$} & \textbf{SR\%} \\
        \midrule
        \multicolumn{13}{l}{\textit{W/ Thinking}} \\
        \hdashline
        Gemini-2.5-pro-think & -0.28 & 81 & 1.91 & 92 & 0.21 & \textbf{84} & \textbf{0.35} & \textbf{74} & -0.04 & 35 & \textbf{0.04} & \textbf{68} \\
        Claude3.7-sonnet-think & \textbf{0.41} & \textbf{81} & 0.59 & 77 & 0.09 & 73 & 0.18 & 66 & -0.01 & 49 & 0.01 & 57 \\
        DeepSeek-R1 & 0.36 & 74 & 1.48 & \textbf{97} & 0.05 & 72 & 0.10 & 62 & -0.06 & 29 & -0.02 & 41 \\
        o3-mini@20250103 & 0.29 & 68 & 1.15 & 85 & 0.17 & \textbf{86} & 0.18 & 69 & -0.08 & 23 & -0.03 & 45 \\
        o1-mini@20240912 & -0.42 & 52 & 1.78 & 95 & 0.07 & 70 & -0.03 & 37 & -0.10 & 15 & -0.08 & 31 \\
        Qwen3-235B-A22B-think & -0.01 & 41 & 0.27 & 42 & 0.01 & 24 & 0.03 & 31 & -0.01 & 23 & 0.01 & 31 \\
        Qwen3-32B-think & 0.0 & 2 & 0.11 & 23 & 0.02 & 14 & 0.0 & 6 & -0.02 & 6 & -0.02 & 5 \\
        Llama-Nemo-49B-think & -0.64 & 24 & 0.20 & 24 & -0.16 & 41 & -0.05 & 30 & -0.15 & 7 & -0.12 & 11 \\
        \midrule
        \multicolumn{13}{l}{\textit{W/o Thinking}} \\
        \hdashline
        GPT-4o@20241120 & -0.20 & 42 & 0.82 & 80 & 0.05 & 70 & 0.05 & 48 & -0.05 & 30 & -0.04 & 39 \\
        Deepseek-V3 & 0.08 & 34 & 0.47 & 93 & 0.08 & 46 & 0.02 & 28 & 0.0 & 18 & 0.0 & 29 \\
        Gemini-2.0-flash & 0.35 & 75 & 0.19 & 54 & 0.10 & 79 & 0.15 & 63 & \textbf{0.03} & 34 & 0.0 & 38 \\
        Qwen235B-A22B & 0.02 & 41 & 0.51 & 45 & 0.01 & 26 & 0.01 & 31 & -0.01 & 23 & 0.0 & 34 \\
        Qwen3-32B & -0.03 & 2 & 0.17 & 23 & 0.02 & 14 & -0.01 & 6 & -0.02 & 6 & -0.02 & 5 \\
        Qwen2.5-72B-Instruct & -0.12 & 42 & 0.28 & 60 & 0.03 & 57 & 0.04 & 40 & -0.02 & 26 & -0.01 & 40 \\
        Qwen2.5-32B-Instruct & 0.03 & 47 & 0.42 & 66 & -0.01 & 54 & 0.04 & 32 & -0.04 & 19 & -0.02 & 31 \\
        Llama-3.1-70B-Instruct & -0.16 & 42 & 0.61 & 80 & 0.07 & 61 & -0.02 & 31 & -0.04 & 30 & -0.02 & 40 \\
        Llama-Nemo-super-49B & -0.14 & 27 & 0.31 & 41 & 0.02 & 50 & -0.02 & 18 & -0.04 & 16 & -0.03 & 27 \\
        Gemma-2-27b-it & -0.03 & 34 & 0.34 & 66 & 0.05 & 56 & -0.03 & 15 & 0.0 & 16 & -0.01 & 17 \\
        Phi-4-14B & -0.10 & 45 & 0.28 & 54 & 0.11 & 74 & -0.04 & 18 & -0.05 & 14 & -0.04 & 22 \\
        OLMo2-32B-Instruct & -2.03 & 22 & 1.03 & 46 & -0.13 & 40 & -0.11 & 7 & -0.12 & 8 & -0.11 & 12 \\
        BioMedGPT-7B & -0.36 & 17 & 0.25 & 63 & -0.29 & 7 & -0.09 & 5 & -0.11 & 6 & -0.08 & 1 \\
        BioMistral-7B & 0.01 & 1 & 0.24 & 6 & 0.0 & 0 & 0.0 & 1 & -0.01 & 1 & -0.01 & 0 \\
        \midrule
        \textbf{MolOptAgent-7B} & \textbf{0.89} & \textbf{92} & \textbf{1.42} & \textbf{84} & \textbf{0.04} & \textbf{35} & \textbf{0.02} & \textbf{38} & \textbf{-0.04} & \textbf{14} & \textbf{0.04} & \textbf{36} \\
        \textbf{MolOptAgent-3B} & \textbf{-0.24} & \textbf{12} & \textbf{-0.021} & \textbf{8} & \textbf{-0.017} & \textbf{5} & \textbf{-0.009} & \textbf{7} & \textbf{-0.003} & \textbf{3} & \textbf{-0.0026} & \textbf{10} \\
        \bottomrule
    \end{tabular}
    }
\end{table}

\noindent
\textbf{MolOptAgent-3B} achieves modest success rates (3--12\% SR) across all objectives, demonstrating that the MolAct framework scales to smaller models, though with reduced performance.

\subsection{Ablation Study}\label{sec:ablation}

\paragraph{Why not just use LLM + tools?}
A natural question is whether instruction-tuned LLMs with tool access can solve molecular optimization through one-stage training (direct optimization) without the two-stage approach (editing pretraining followed by optimization). We train Qwen-2.5-3B/7B-Instruct models using agentic RL on optimization tasks only (one-stage), with the same tools and interaction budget (\texttt{max\_turns = 16}) as MolOptAgent. As shown in Table~\ref{tab:ablation_molopt_sr}, both one-stage models yield near-zero success rates across all objectives (Qwen-2.5-7B: 12\% on QED only; Qwen-2.5-3B: 0\% across all tasks). This failure occurs despite having tool access and RL training. The bottleneck is not molecular knowledge or reasoning capacity, but the absence of learned policies for tool usage and termination.

\begin{figure}[!h]
    \centering
    \begin{subfigure}{0.48\linewidth}
        \centering
        \includegraphics[width=\linewidth]{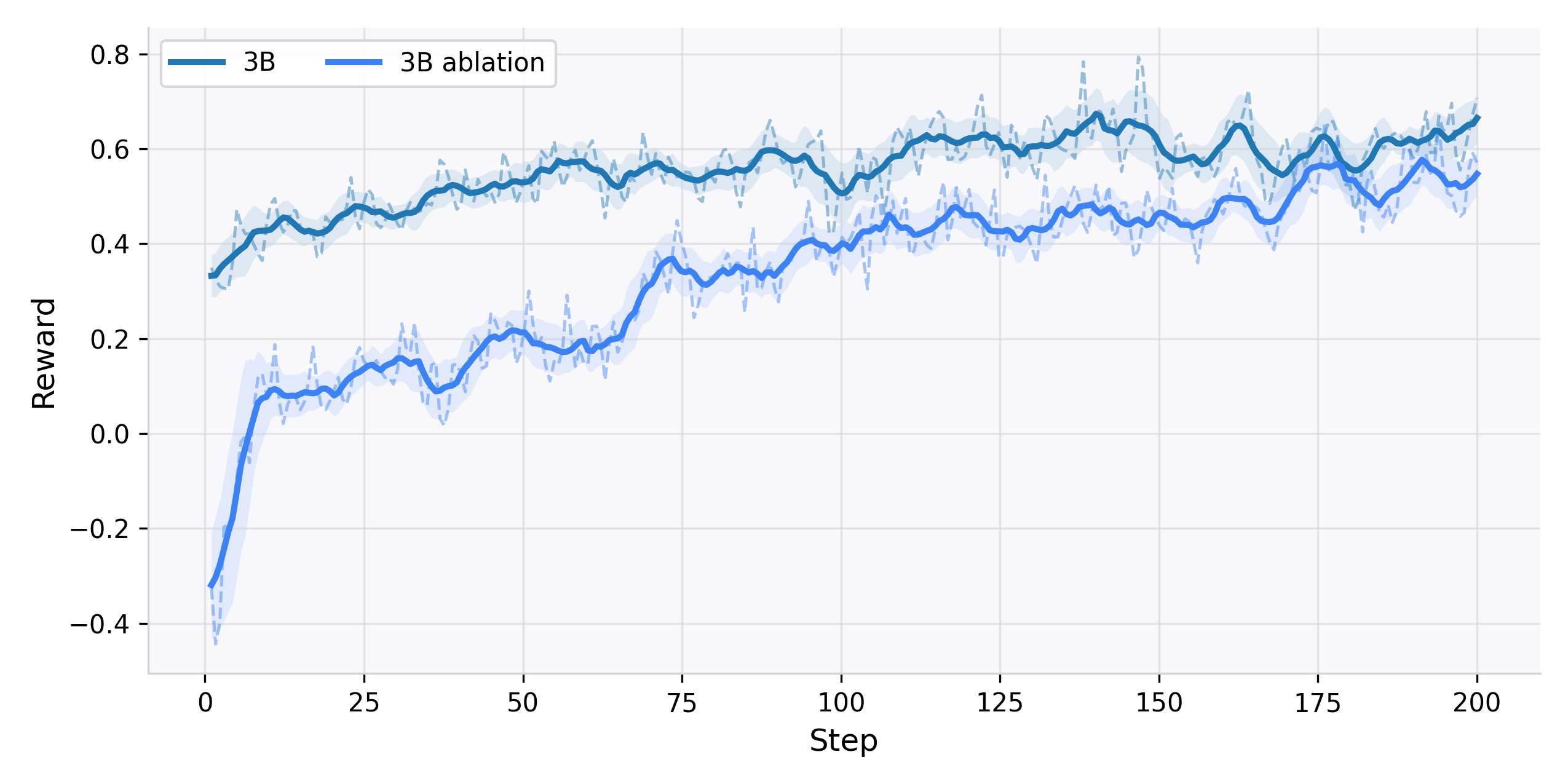}
        \caption{3B backbone (Qwen-2.5-3B): two-stage pretraining on editing accelerates reward growth and lifts the plateau.}
        \label{fig:ablation_3b}
    \end{subfigure}
    \hfill
    \begin{subfigure}{0.48\linewidth}
        \centering
        \includegraphics[width=\linewidth]{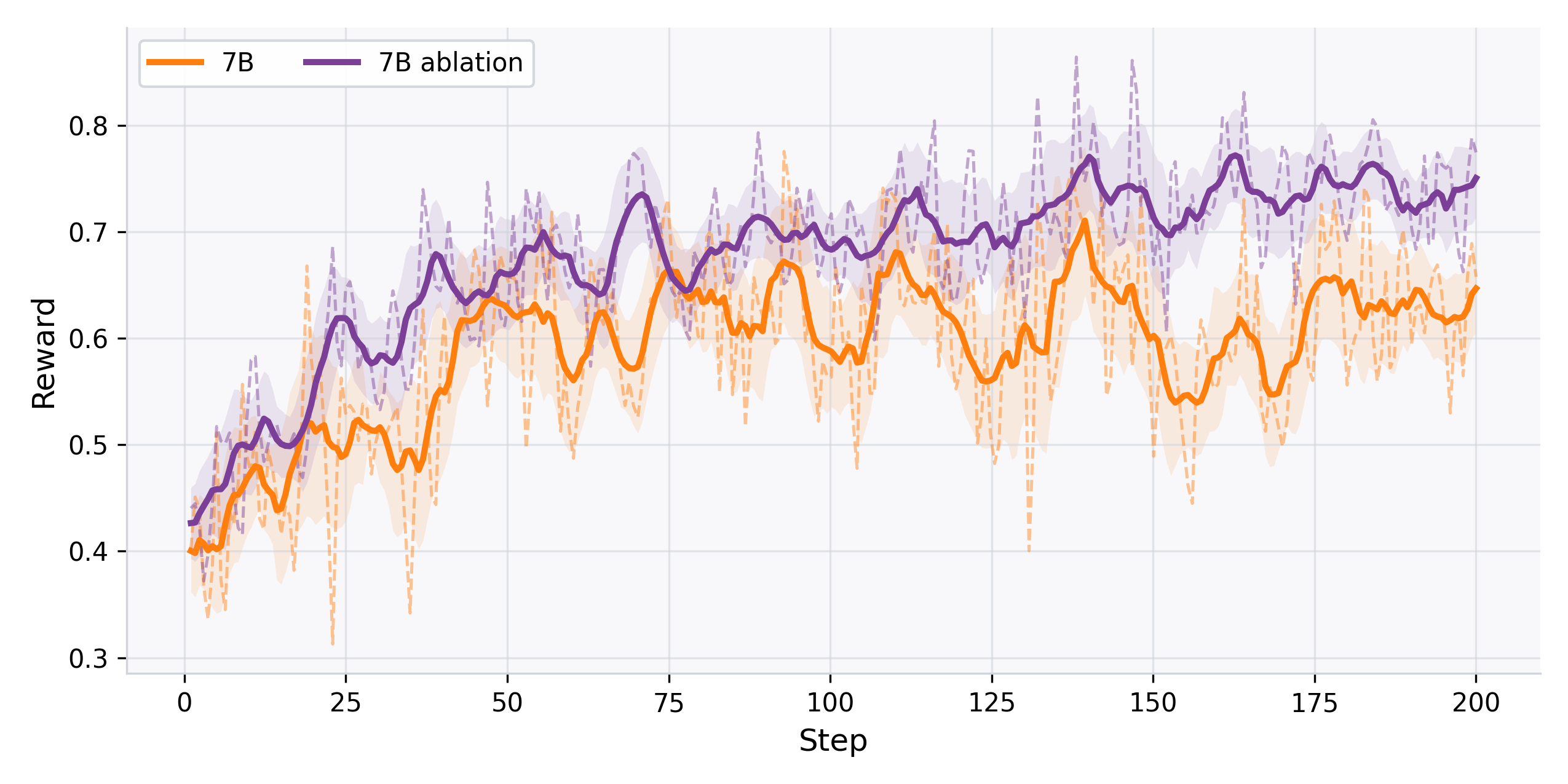}
        \caption{7B backbone (Qwen-2.5-7B): two-stage training improves convergence stability and final reward.}
        \label{fig:ablation_7b}
    \end{subfigure}
    \caption{One-stage vs. two-stage training on molecular optimization across backbone scales.}
    \label{fig:ablation_combo}
\end{figure}

\paragraph{Why does two-stage training work?}
Two-stage training (editing pretraining followed by optimization) provides essential foundation for learning tool usage and termination policies. Figure~\ref{fig:ablation_combo} illustrates that two-stage training accelerates reward convergence and lifts final rewards compared to one-stage training for both 3B and 7B backbones. The performance gap is reflected in success rates: as shown in Table~\ref{tab:ablation_molopt_sr}, MolOptAgent-7B (two-stage) achieves far higher success rates (LogP 92\%, Solubility 84\%, QED 35\%, DRD2 38\%, JNK3 14\%, GSK-3$\beta$ 36\%) compared to Qwen-2.5-7B-Instruct (one-stage). Similarly, MolOptAgent-3B (two-stage) achieves 3--12\% SR across tasks, while Qwen-2.5-3B-Instruct (one-stage) achieves 0\% on all tasks. This confirms that editing pretraining is critical for learning effective tool usage and termination policies before tackling optimization tasks.

\begin{table}[!h]
    \centering
    \caption{Ablation: success rate (SR\%) comparison under the same interaction budget. One-stage training (optimization only) fails to learn effective tool usage and termination policies, while two-stage training (editing then optimization) achieves high success rates. Column order follows Table~\ref{tab:molopt_performance}.}
    \label{tab:ablation_molopt_sr}
    \begin{tabular}{l|cccccc}
        \toprule
        \textbf{Model} & \textbf{LogP} & \textbf{Solubility} & \textbf{QED} & \textbf{DRD2} & \textbf{JNK3} & \textbf{GSK-3$\beta$} \\
        \midrule
        \multicolumn{7}{l}{\textit{One-stage training (optimization only)}} \\
        \hdashline
        Qwen-2.5-7B-Instruct & 0 & 0 & 12 & 0 & 0 & 0 \\
        Qwen-2.5-3B-Instruct & 0 & 0 & 0 & 0 & 0 & 0 \\
        \midrule
        \multicolumn{7}{l}{\textit{Two-stage training (editing then optimization)}} \\
        \hdashline
        \textbf{MolOptAgent-7B} & \textbf{92} & \textbf{84} & \textbf{35} & \textbf{38} & \textbf{14} & \textbf{36} \\
        \textbf{MolOptAgent-3B} & \textbf{12} & \textbf{8} & \textbf{5} & \textbf{7} & \textbf{3} & \textbf{10} \\
        \bottomrule
    \end{tabular}
\end{table}

\begin{figure*}[!h]
    \centering
    \begin{subfigure}{0.48\linewidth}
        \centering
        \includegraphics[width=\linewidth]{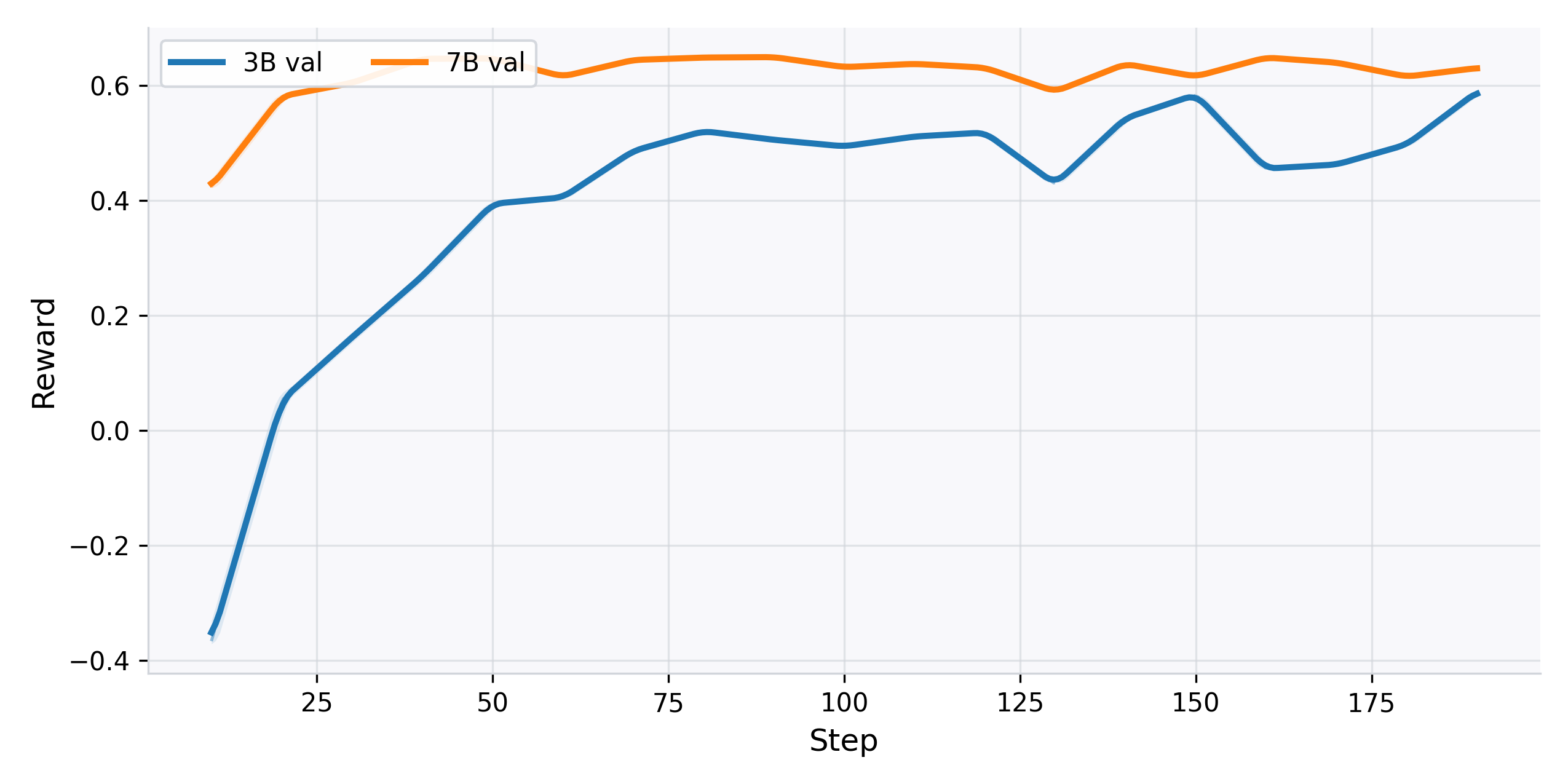}
        \caption{Molecular editing: MolEditAgent-3B vs. MolEditAgent-7B. The 7B variant converges faster and to higher rewards.}
        \label{fig:mol_edit_val}
    \end{subfigure}
    \hfill
    \begin{subfigure}{0.48\linewidth}
        \centering
        \includegraphics[width=\linewidth]{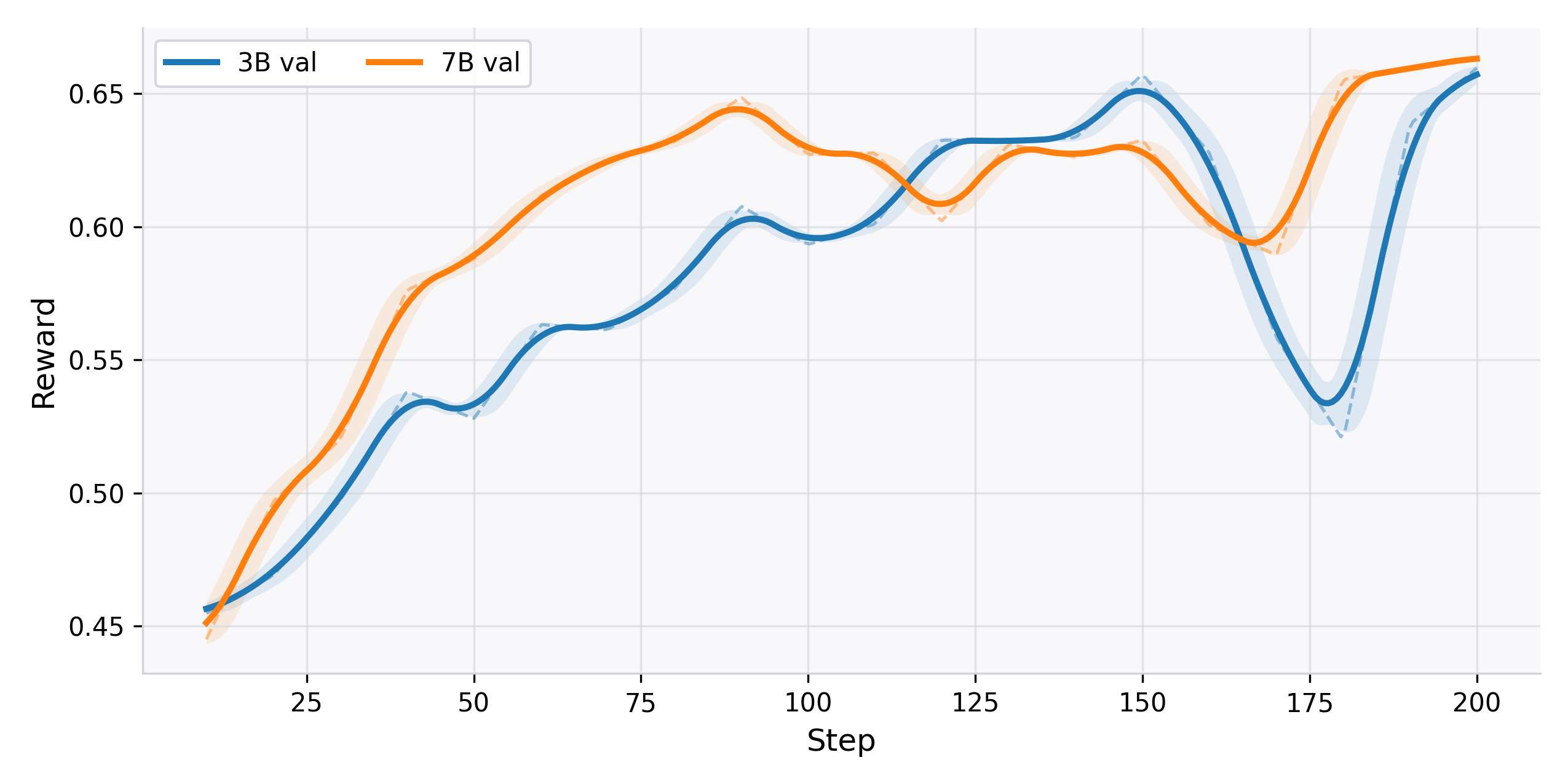}
        \caption{Molecular optimization: MolOptAgent-3B vs. MolOptAgent-7B. Rewards converge similarly, but success rates diverge (3--12\% vs. 14--92\%).}
        \label{fig:mol_opt_val}
    \end{subfigure}
    \caption{Validation reward trajectories for 3B vs. 7B. The 7B models converge faster; in optimization, similar rewards mask large success-rate gaps, underscoring the role of capacity in executing tool-augmented policies.}
    \label{fig:val_reward_combo}
\end{figure*}

\paragraph{Model capacity and tool usage efficiency.}
Figure~\ref{fig:val_reward_combo} compares validation rewards for editing and optimization. In editing, MolEditAgent-7B attains higher rewards and matches its higher accuracy. In optimization, MolOptAgent-3B and MolOptAgent-7B reach similar rewards, yet success rates diverge (3--12\% vs. 14--92\%) under the same \texttt{max\_turns = 16} budget. The 3B model struggles to sequence tools and terminate correctly, so reward signals do not translate to task success; the 7B model executes learned policies reliably.

\begin{figure}[!h]
    \centering
    \begin{subfigure}{0.48\linewidth}
        \centering
        \includegraphics[width=\linewidth]{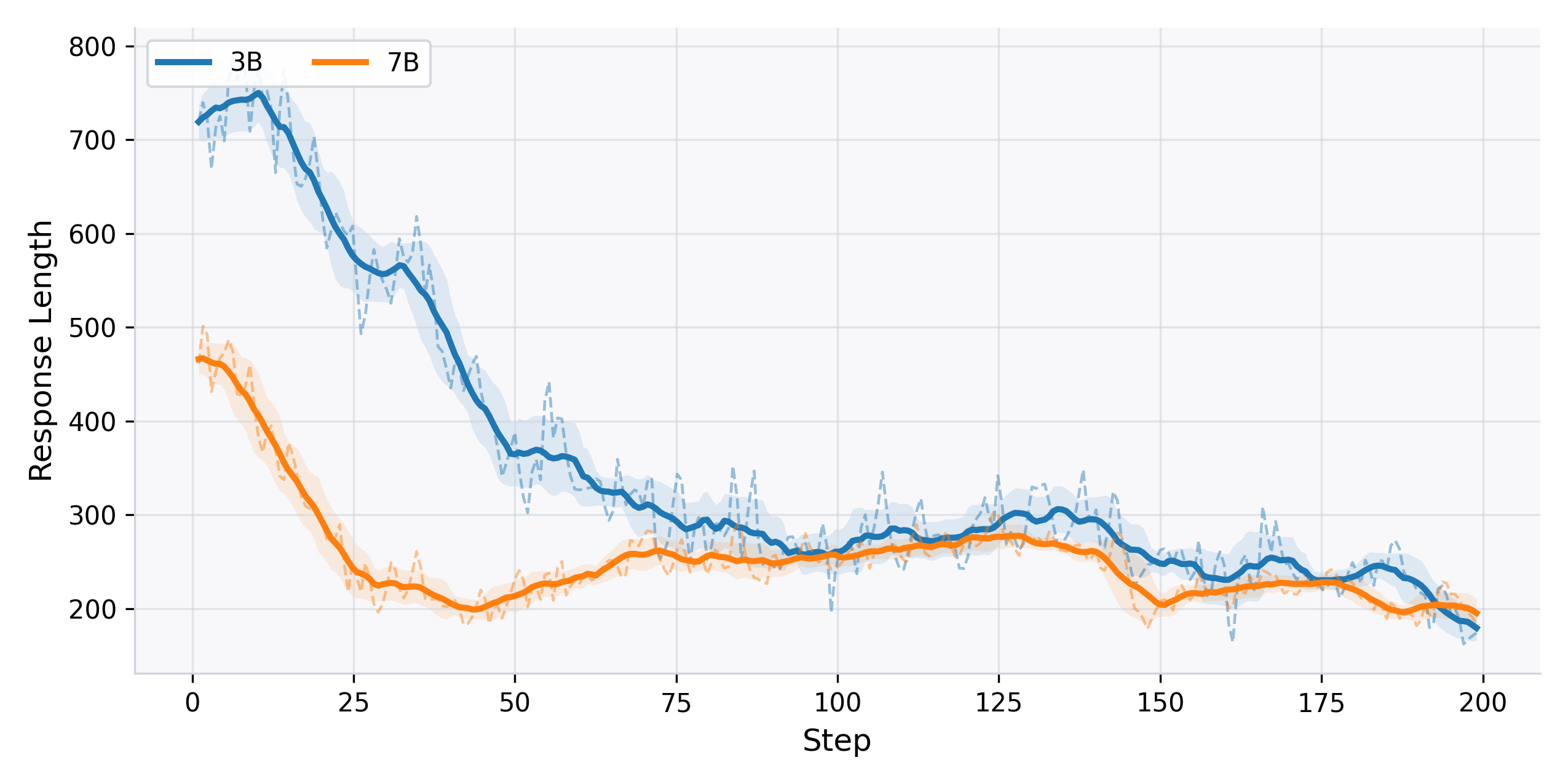}
        \caption{Molecular editing: average response length.}
        \label{fig:response_length_edit}
    \end{subfigure}
    \hfill
    \begin{subfigure}{0.48\linewidth}
        \centering
        \includegraphics[width=\linewidth]{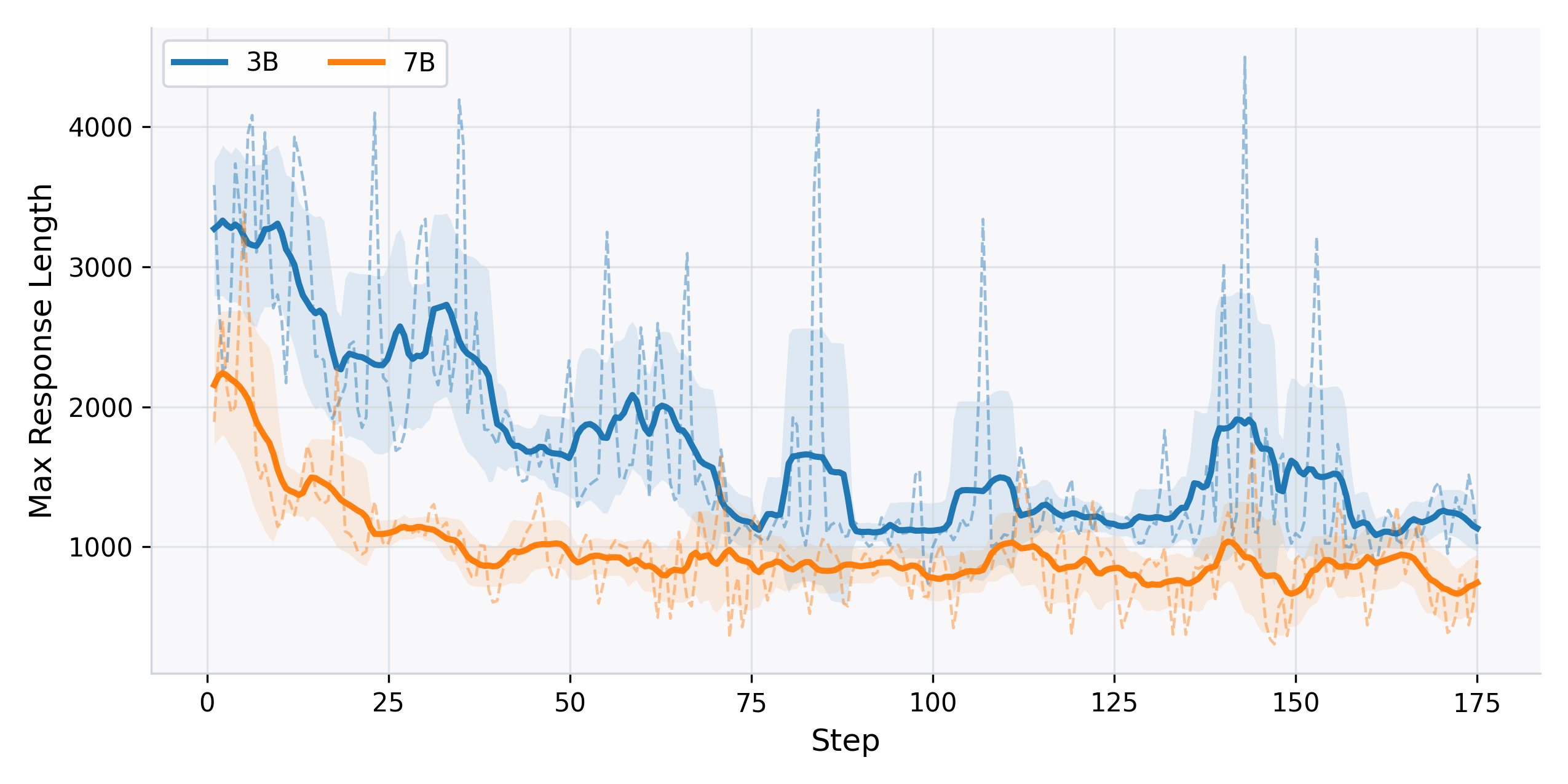}
        \caption{Molecular editing: max response length.}
        \label{fig:response_length_edit_max}
    \end{subfigure}
    \caption{Response lengths during MolEditAgent training. Lower, more stable lengths for 7B indicate more efficient tool use.}
    \label{fig:response_length_edit_combo}
\end{figure}

\begin{figure}[!h]
    \centering
    \begin{subfigure}{0.48\linewidth}
        \centering
        \includegraphics[width=\linewidth]{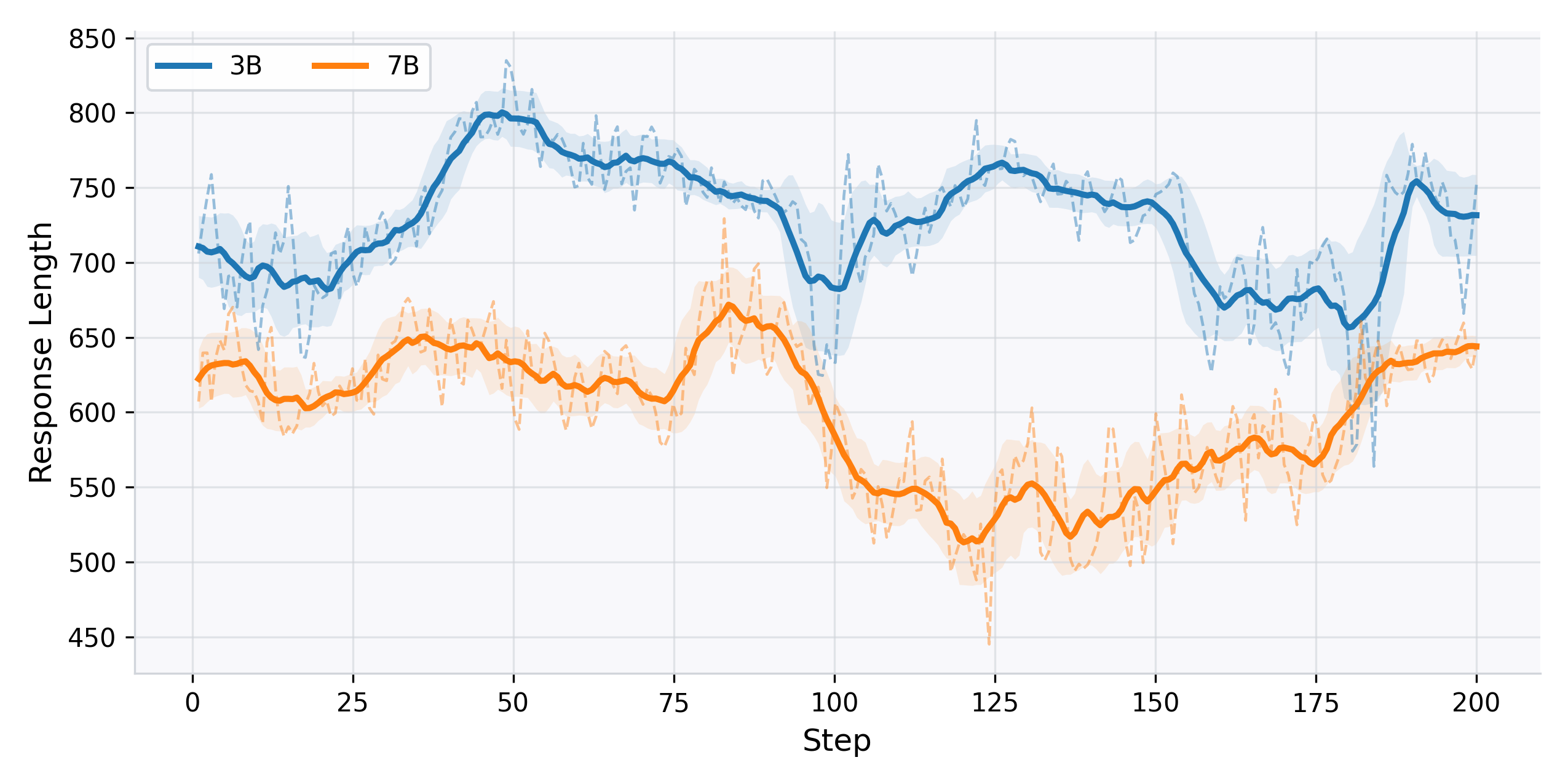}
        \caption{Molecular optimization: average response length.}
        \label{fig:response_length_opt}
    \end{subfigure}
    \hfill
    \begin{subfigure}{0.48\linewidth}
        \centering
        \includegraphics[width=\linewidth]{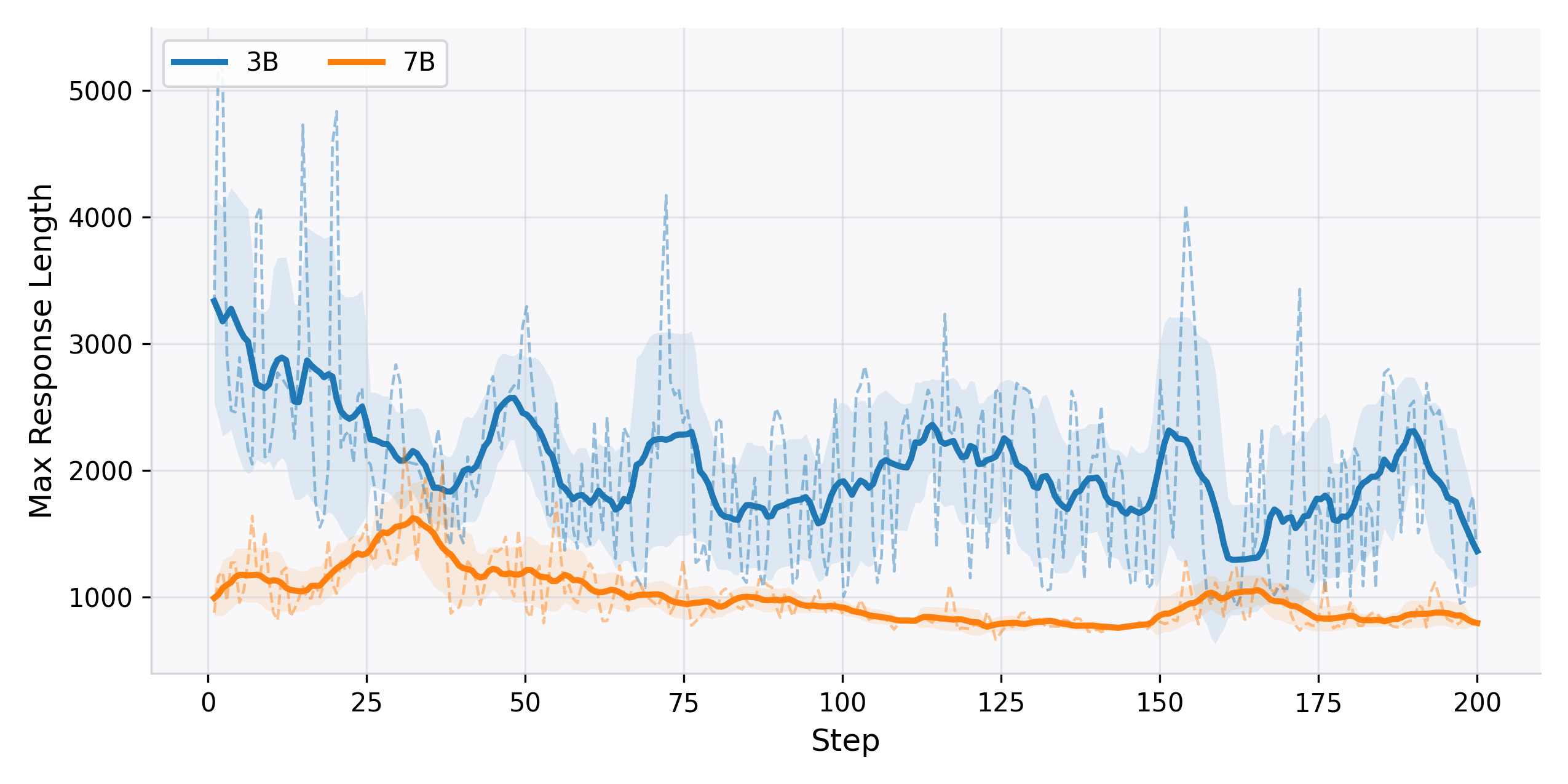}
        \caption{Molecular optimization: max response length.}
        \label{fig:response_length_opt_max}
    \end{subfigure}
    \caption{Response lengths during MolOptAgent training. The 3B model shows extreme spikes (up to 5{,}000 tokens), indicating inefficient tool use; the 7B model remains shorter and steadier.}
    \label{fig:response_length_opt_combo}
\end{figure}

To probe this gap, we analyze response length as a proxy for tool usage efficiency. Efficient agents issue short, focused responses with appropriate tool calls; unstable or verbose responses indicate poor tool sequencing. In editing (figures~\ref{fig:response_length_edit} and~\ref{fig:response_length_edit_max}), MolEditAgent-7B maintains low, stable lengths, while MolEditAgent-3B is more variable. In optimization (figures~\ref{fig:response_length_opt} and~\ref{fig:response_length_opt_max}), MolOptAgent-3B exhibits large spikes in maximum length (up to 5{,}000 tokens), reflecting verbose, inefficient reasoning; MolOptAgent-7B keeps shorter, steadier responses. This explains why similar rewards yield very different success rates: smaller models fail to execute tool-augmented policies within the interaction budget.

\section{Discussion}

In this work we view molecular editing and optimization as multi-step, tool-augmented decisions and instantiate MolAct to train MolEditAgent and MolOptAgent. Across benchmarks, iterative control with explicit tool feedback yields high validity and competitive property outcomes, and larger backbones translate learned rewards into executable policies more reliably.

Several lessons emerge. First, curriculum matters: mastering edit operations before property optimization is critical under tight turn budgets. Second, capacity matters: smaller models can fit rewards yet fail to execute tool sequences and terminate on budget, so executability must be evaluated beyond reward metrics. Third, domain specificity matters: targets like JNK3 demand specialized knowledge and tools that general-purpose backbones lack without augmentation. These observations argue for explicit tool curricula, executability-aware evaluation, and domain-tailored extensions when building agentic molecular systems.

From a molecular design perspective, our work demonstrates that multi-step, tool-guided processes can effectively bridge the gap between molecular reasoning and actionable modifications.
The success of MolAct in maintaining high validity (95--100\% for 7B) while achieving competitive optimization performance suggests that explicit verification and feedback loops are essential for reliable molecular design, aligning with real-world medicinal chemistry practice where iterative refinement with intermediate validation is standard.

Despite encouraging results, several limitations remain. First, the approach depends on external property oracles; bias or inaccuracy in these oracles can misguide optimization trajectories \citep{Renz2019-ja,gao2022sampleefficiencymattersbenchmark}. Second, predefined edit operators and a fixed interaction budget (\texttt{max\_turns = 16}) restrict exploration of more complex transformations. Third, performance on certain specific objectives (e.g., JNK3) remains modest; these targets demand specialized knowledge and tools that small backbones (3B/7B)—and even many larger general models—lack without domain-tailored augmentation. Fourth, while two-stage training is effective, it presumes transfer from editing to optimization; some property–objective pairs may need tailored curricula or tools. Finally, we do not model synthetic feasibility or reaction pathways \citep{Gao2020-ak, Polykovskiy2020-ox}, so some generated molecules may be hard to synthesize despite being valid in silico.

There are several promising directions for future research.
One important direction is to incorporate synthetic feasibility and reaction-aware constraints during molecule optimization, bridging the gap between theoretical optimization and practical synthesis.
Another potential extension is to explore more sophisticated error recovery mechanisms and adaptive interaction strategies to improve smaller models' ability to execute learned policies.
Finally, investigating curriculum learning strategies beyond two-stage training and developing evaluation metrics that assess policy executability alongside reward learning could further enhance the effectiveness of agentic molecular design systems.
Building on MolAct, integrating diverse RL algorithms, richer toolsets, and tailored reward designs within the same multi-turn framework could yield more sample-efficient training and broader task coverage.

\section{Method}

This section describes how the MolAct framework instantiates the sequential decision process as a tool-augmented, group-relative policy optimization framework for molecular editing and optimization. The framework can be used to train different model families: MolEditAgent for editing tasks and MolOptAgent for optimization tasks. We first outline the rollout and optimization scheme, then describe the tool interface, and finally detail the reward definitions.

\subsection{Group-Relative Tool-Augmented Optimization}
For each prompt (source SMILES and task specification), we copy it into $K$ parallel rollout chains that form one group. Each chain is a multi-turn sequence of agent responses and tool observations; chains in the same group share the same prompt so their rewards can be normalized together, following GRPO~\citep{shao2024deepseekmathpushinglimitsmathematical}. A rollout stops when the agent emits \texttt{terminate} or hits a turn budget. The full token sequence concatenates prompt, all responses, and tool outputs; we apply a binary mask over tokens so that gradients and advantages are computed only on agent-generated tokens, treating tool outputs and prompts as fixed context.

Concretely, for a trajectory with $L$ tokens $\{a_t\}_{t=1}^L$ and mask $M_t \in\{0,1\}$ indicating agent tokens, the surrogate loss is
\[
\mathcal{L}(\theta) = \sum_{t=1}^L M_t \,\ell\big(a_t, s_t, \widehat{A}_t\big),
\]
where $s_t$ is the prefix up to $t$ and $\widehat{A}_t$ is the advantage estimate. Advantages are computed group-relatively within the $K$ rollouts of the same prompt to reduce variance in long-horizon, tool-mediated interactions. This “group-relative, masked-token” optimization stabilizes credit assignment while preserving full tool feedback in context.

\subsection{Tool-Augmented Molecular Agent}
The agent can (i) invoke edit operators (add, delete, substitute) with tool-validated sites, (ii) query validity, similarity, or property oracles (LogP, solubility, QED, DRD2, JNK3, GSK3$\beta$), or (iii) terminate. Tool calls return observations that stay in context but do not receive gradients. A typical reasoning trace during training or inference follows: prompt $\rightarrow$ \texttt{think} (propose) $\rightarrow$ tool call $\rightarrow$ observation $\rightarrow$ \texttt{think} $\rightarrow$ tool call $\rightarrow \dots \rightarrow$ \texttt{terminate}, after which the final SMILES is scored and, during training, its reward is propagated via the masked-token loss.

\subsection{Reward Design}
Let $s_{\mathrm{pred}}$ be the final SMILES and $s_{\mathrm{src}}$ the input. Rewards are episode-level, clipped to $[-1,1]$, and applied only if $s_{\mathrm{pred}}$ is valid:
\[
r = 
\begin{cases}
-1, & \text{if } s_{\mathrm{pred}} \text{ is invalid},\\
\mathrm{clip}_{[-1,1]}\big(0.8\, r_{\mathrm{task}} + 0.15\, r_{\mathrm{struct}} + 0.05\, r_{\mathrm{tool}}\big), & \text{otherwise}.
\end{cases}
\]
\paragraph{Editing.} Task term $r_{\mathrm{task}}$ is binary correctness of the instructed operator. For a target group $\mathcal{G}$ and counts $c(s,\mathcal{G})$:
\[
r_{\mathrm{task}} =
\begin{cases}
\mathbb{I}[c(s_{\mathrm{pred}},\mathcal{G}) = c(s_{\mathrm{src}},\mathcal{G}) + 1], & \text{add},\\
\mathbb{I}[c(s_{\mathrm{pred}},\mathcal{G}) = c(s_{\mathrm{src}},\mathcal{G}) - 1], & \text{delete},\\
\mathbb{I}[c(s_{\mathrm{pred}},\mathcal{G}_{\mathrm{add}})=c(s_{\mathrm{src}},\mathcal{G}_{\mathrm{add}})+1 \wedge c(s_{\mathrm{pred}},\mathcal{G}_{\mathrm{del}})=c(s_{\mathrm{src}},\mathcal{G}_{\mathrm{del}})-1], & \text{substitute}.
\end{cases}
\]
Structural term $r_{\mathrm{struct}} = \max\{0, \mathrm{Sim}_{\mathrm{tan}}(s_{\mathrm{pred}}, s_{\mathrm{ref}})\}$ uses a reference SMILES $s_{\mathrm{ref}}$ (post-edit target or task reference). Tool bonus $r_{\mathrm{tool}}=\mathbb{I}_{\mathrm{tool}}$ encourages grounded calls.

\paragraph{Optimization.} Task term $r_{\mathrm{task}}$ is normalized property improvement. For target oracle $p(\cdot)$ and threshold $\delta$:
\[
r_{\mathrm{task}} = \mathrm{clip}_{[0,1]}\!\left(\frac{p(s_{\mathrm{pred}})-p(s_{\mathrm{src}})}{\delta}\right),
\]
with $\delta=0.5$ for LogP, solubility, and QED, and $\delta=0.3$ for DRD2, JNK3, and GSK3$\beta$. Structural term $r_{\mathrm{struct}}=\mathrm{Sim}_{\mathrm{scaf}}(s_{\mathrm{pred}}, s_{\mathrm{src}})$ uses Murcko scaffold Tanimoto; tool bonus follows editing. This provides a compact, bounded signal for long-horizon rollouts.

\section{Related Work}


\subsection{Molecular Editing and Optimization}

Early work on molecular editing and optimization primarily relied on
SMILES-based or graph-based generative models, including variational autoencoders \citep{DBLP:journals/corr/abs-1802-04364,DBLP:journals/corr/abs-1802-03480},
autoregressive generators \citep{Gomez-Bombarelli2018-ly, jin2019hierarchicalgraphtographtranslationmolecules, fang2024domainagnosticmoleculargenerationchemical}, and search-based methods \citep{Jensen2019-vj, Zhang2025.08.19.671029, Zhang2025-hd}.
Representative approaches optimize molecular properties by directly generating
candidate molecules or by exploring the chemical space through heuristic-guided search.
While effective in specific settings, these methods often depend on handcrafted
model architectures or predefined optimization pipelines, which limits their adaptability
to diverse objectives and complex structural constraints.

More recently, large language models have been applied to molecular design tasks,
leveraging their strong representation learning and instruction-following capabilities \citep{ye2023drugassistlargelanguagemodel, Lei2025MolEditKE}.
These methods typically treat molecule modification as a static generation problem
or as instruction-driven editing, where a modified molecule is produced in a single step.
Although promising, such approaches do not explicitly model the sequential nature
of molecular editing and optimization and lack mechanisms for incorporating
structured feedback from intermediate molecular evaluations.

\subsection{Agentic Reinforcement Learning for LLMs}

Language model agents have emerged as a paradigm for enabling language models
to interact with external tools and environments \citep{dong2025toolstarempoweringllmbrainedmultitool}.
Early agent systems were primarily constructed through prompt engineering
or supervised fine-tuning on curated interaction trajectories.
Subsequent work has explored reinforcement learning as a means
to improve long-horizon reasoning and exploration capabilities
of language model agents \citep{wang2025agentflyextensiblescalablereinforcement, xi2025agentgymrltrainingllmagents, fu2025areallargescaleasynchronousreinforcement}.

Agentic reinforcement learning introduces additional challenges,
including multi-turn credit assignment, scalable rollout,
and the integration of heterogeneous tools \citep{gao2025txagentaiagenttherapeutic, gao2025democratizingaiscientistsusing}.
Recent frameworks address these challenges by unifying tool invocation
and environment interaction within reinforcement learning pipelines,
making it possible to train agents that operate over extended interaction horizons.
These advances provide a foundation for applying agentic reinforcement learning
to complex, real-world decision-making problems, such as drug discovery, materials science, and robotics control, which extend far beyond traditional language tasks.

\newpage

\bibliography{main}
\bibliographystyle{colm2025_conference}

\newpage







\end{CJK*}
\end{document}